\documentclass[preprint,12pt]{elsarticle}

\usepackage{graphicx}
\usepackage{subfig}

\usepackage{amsmath}
\usepackage{lineno}
\usepackage[colorlinks,linkcolor=blue]{hyperref}

\journal{Image and Vision Computing}

\begin{document}

\begin{frontmatter}


\title{IoU-aware Single-stage Object Detector for Accurate Localization}

%% \title{Titlee\tnoteref{label1}}
% \author{Name\corref{cor1}\fnref{label2}}
% \address{Address\fnref{label3}}

% \author[label1,label2]{<author name>}
% \address[label1]{<address>}
% \address[label2]{<address>}
\author[mymainaddress]{Shengkai Wu}
\ead{ShengkaiWu@hust.edu.cn}
\author[mymainaddress]{Xiaoping Li\corref{mycorrespondingauthor}}
\cortext[mycorrespondingauthor]{Corresponding author}
\ead{lixiaoping@hust.edu.cn}
\author[mysecondaryaddress]{Xinggang Wang}
\ead{xgwang@hust.edu.cn}
\address[mymainaddress]{State Key Laboratory of Digital Manufacturing Equipment and Technology, Huazhong University of Science and Technology, Wuhan, 430074, China.}
\address[mysecondaryaddress]{School of EIC, Huazhong University of Science and Technology, Wuhan, 430074, China.}

\begin{abstract}
Single-stage object detectors have been widely applied in many computer vision applications due to their simpleness and high efficiency. However, the low correlation between the classification score and localization accuracy in detection results severely hurts the average precision of the detection model. To solve this problem, an IoU-aware single-stage object detector is proposed in this paper. Specifically, IoU-aware single-stage object detector predicts the IoU for each detected box. Then the predicted IoU is multiplied by the classification score to compute the final detection confidence, which is more correlated with the localization accuracy. The detection confidence is then used  as the input of the subsequent NMS and COCO AP computation, which substantially improves the localization accuracy of model. Sufficient experiments on COCO and PASCOL VOC datasets demonstrate the effectiveness of IoU-aware single-stage object detector on improving model's localization accuracy. Without whistles and bells, the proposed method can substantially improve AP by $1.7\%\sim1.9\%$ and AP75 by $2.2\%\sim2.5\%$ on COCO \textit{test-dev}. And it can also substantially improve AP by $2.9\%\sim4.4\%$ and AP80, AP90 by $4.6\%\sim10.2\%$ on PASCAL VOC. Code is available \href{https://github.com/ShengkaiWu/IoU-aware-single-stage-object-detector}{here}.

\end{abstract}


\end{frontmatter}

% \linenumbers

\section{Introduction}
\label{S:1}
As the development of deep convolutional neural networks, a large amount of object detection models have been proposed in recent years. Most of these models can be classified into single-stage  object detectors \cite{liu2016ssd,redmon2016you,lin2017focal,zhang2018single_RefineDet,zhang2018single_Entiched,li2019gradient} and multi-stage object detectors \cite{ren2015faster,cai2018cascade,he2017mask,lin2017FPN,dai2016R-FCN,girshick2015fast,girshick2014r-cnn}. For the multi-stage object detectors, multi-stage classification and localization are applied sequentially, which make these models more powerful on classification and localization tasks. Compared with the single-stage object detectors, the multi-stage object detectors have achieved better average precision(AP), but their efficiency is hurt by the subnetworks of the multi-stage classification and localization. On the contrary, the single-stage detectors rely on a single fully convolutional networks(FCN) for classification and localization, which is more simple and efficient. However, the AP of single-stage detectors generally lag behind that of the multi-stage detectors.

\begin{figure}[h]
\centering
\includegraphics[height=0.34\linewidth]{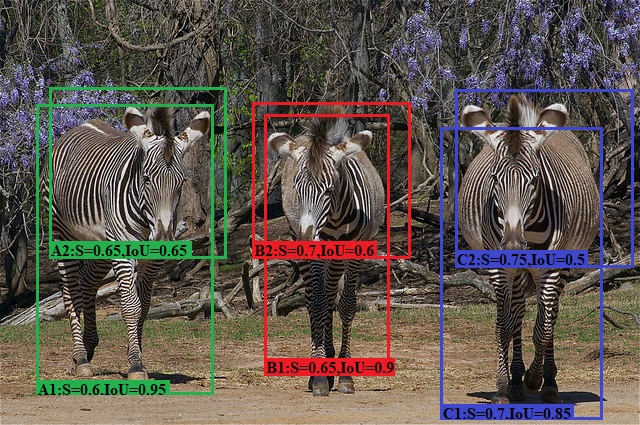}
\caption{The detections having the mismatch problem between the classification score and localization accuracy. "S" represents the classification score. These detections hurt the model's performance in two ways. Firstly, the accurate boxes A1, B1 and C1 are suppressed by inaccurate boxes A2, B2, and C2 respectively during NMS. Secondly, during computing average precision(AP), the survived boxes are ranked as C2, B2, A2, which results in lower AP than that they are ranked as A2 B2, C2.}
\label{fig:mismatchProblem}
\end{figure}

In this work, we aim to improve the AP of single-stage detectors especially the localization accuracy while keeping their efficiency. We demonstrate that the low correlation between the classification score and localization accuracy of single-stage detectors severely hurt the localization accuracy of the models. The low correlation is mostly caused by that the subnetworks of the classification and localization are trained with independent objective functions without knowing each other explicitly. After the models are converged, the classification subnetwork predicts the classification score for each regressed anchor without knowing the localization accuracy, represented by IoU between the regressed anchor and the ground truth box. Thus, there will be many detections having the mismatch problem between the classification scores and their localization accuracy, such as detections with high classification scores but low IoU, detections with low classification scores but high IoU as shown in Fig.\ref{fig:mismatchProblem}. These detections hurt the average precision of the models in two ways during inference. Firstly, during standard non-maximum suppression(NMS), all the detections are ranked based on their classification scores and the detection with the highest classification score will suppress the other detections that have an overlap higher than the manually set threshold with it. Consequently, the detections with low classification scores but high IoU will be suppressed by the detections with high classification scores but low IoU. For example, the accurate boxes A1, B1 and C1 are suppressed by the inaccurate boxes A2, B2 and C2 respectively as shown in Fig.\ref{fig:mismatchProblem}. Secondly, during computing the average precision(AP), all the detections are also ranked according to their classification scores. The precisions and recalls are computed based on these ranked detections and if the detections with high classification scores but low IoU rank before the detections with low classification scores but high IoU, the precision at high IoU threshold will be reduced, which results in lower AP at high IoU threshold. For example, the survived boxes in Fig.\ref{fig:mismatchProblem} are ranked as C2, B2, A2, which results in lower AP than that they are ranked as A2, B2, C2. Both of these hurt the average precision of models.

To solve the above problem, we propose an IoU-aware single-stage object detector based on RetinaNet \cite{lin2017focal}. An IoU prediction head  parallel with the regression head is attached to the last layer of the regression branch to predict the IoU of each regressed anchor. During training, the IoU prediction head is trained jointly with the classification head and regression head. During inference, the detection confidence is computed by multiplying the classification score and predicted IoU for each detected box and then used to rank all the detections in the subsequent NMS and AP computation. Because the detection confidence is more correlated with the localization accuracy, the problem mentioned above can be alleviated and thus the localization accuracy of models can be substantially improved as the experiments show.

The contributions of our paper are as follows: (1) A novel IoU-aware single-stage object detector is proposed to solve the mismatch problem between the classification score and localization accuracy of the single-stage object detector. The method is extremely simple and elegant while the model's performance especially the localization accuracy can be substantially improved without sacrificing the efficiency. (2) We conduct extensive experiments to dive deep into the IoU prediction problem and systemically analyze why IoU-aware single-stage detector works, the performance's bound of our method and the existing gap between the predicted IoU and ground truth IoU. These analyses present a meaningful and inspiring question that which factors are important for the accurate IoU prediction and will inspire the following research.

The rest of this paper is organized as follows. Section \ref{S:2} introduces the related research work. Section \ref{S:3} introduces the IoU-aware single-stage object detector in details. Section \ref{S:4} presents extensive experiments on COCO and PASCAL VOC dataset to demonstrate the effectiveness of our method and some discussions are presented to inspire the following research. Section \ref{S:5} gives the conclusions.

\section{Related Work}
\label{S:2}

\textbf{Correlation between classification score and localization accuracy}. The low correlation between the classification score and localization accuracy hurts the models' localization accuracy severely and many methods have been proposed to solve this problem. Fitness NMS \cite{tychsen2018fitnessNMS} improves DeNet  \cite{tychsen2017denet} by dividing the localization accuracy into 5 levels and transforming the localization accuracy prediction task to the classification task. During inference, the fitness for each detected box is computed as the weighted sum of the predicted fitness probabilities and then multiplied by the classification score as the final detection score which is more correlated with the localization accuracy. Then the final detection score is used as the input of NMS, denoted as Fitness NMS, to improve the localization accuracy of DeNet. IoU-Net \cite{jiang2018IoUNet} improves Faster R-CNN \cite{ren2015faster} by designing an IoU prediction head parallel with the R-CNN to predict the regressed IoU for each RoI. During inference, all the detected boxes are ranked based on the predicted IoU and then IoU-guided NMS is applied to improve the localization accuracy. Compared with IoU-Net, the IoU prediction head in our method is extremely light and the IoU-guided NMS is not used. Similarly, MS R-CNN \cite{huang2019MSR-CNN} improves Mask R-CNN \cite{he2017mask} by attaching a MaskIoU head parallel with the Mask head to predict the IoU between the predicted mask and the corresponding ground truth mask. During inference, the predicted IoU is multiplied by the classification score as the final mask confidence used to rank the predicted mask when computing AP. YOLO \cite{redmon2016you} proposes a real-time single-stage object detector and IoU is also predicted to encode the information how well the predicted box fits the object. Precise detection \cite{goldman2019precise} aims to solve the object detection in the man-made scenes such as retail shelf where objects are densely packed and positioned in close proximity. A Soft-IoU layer is designed to predict the quality scores and then the quality score is used in the proposed EM merging unit to resolve detection overlap ambiguities. All the above methods design additional subnetworks to predict the localization accuracy. There also exists other research solving the problem by designing better loss functions without changing the models' architecture. PISA \cite{cao2019PISA} assigns different weights to the positive examples in the classification loss based on their importance which is obtained by IoU Hierarchical Local Rank (IoU-HLR). In addition, the classification probabilities are used to reweight the contribution of each positive example to the regression loss, denoted as classification-aware regression loss. Both the improvements to the classification and regression loss can enhance the correlation between the classification score and localization accuracy. Similarly, the IoU-balanced classification loss \cite{2019IoU-balancedlosses} uses the regressed IoU to reweight the classification loss for each positive example directly and aims to make the examples with higher IoU learn higher classification score, which thus enhances the correlation between classification score and localization accuracy. The IoU-aware single-stage object detector aims to improve RetinaNet with minimum modification to the original model. A single IoU prediction layer is added in the regression branch to predict the IoU for each detection, which adds negligible computation burden. The predicted IoU is multiplied by the classification score as the final detection confidence which is directly used in the subsequent NMS and AP computation procedure. This minor modification can bring substantially improvement to the model's performance without sacrificing the efficiency.%\footnote{\hl{Add some description on the difference between this paper and IoU-Net.}}

\textbf{Accurate object localization}. Accurate object localization is extremely challenging in the complex scene such as COCO dataset and a large number of methods have been proposed to improve the localization accuracy of object detection models in recent years. Multi-region detector \cite{gidaris2015multiregion} finds that a single-stage regression is limited for accurate localization and thus an iterative bounding box regression procedure is proposed to refine the coordinates of detected boxes, followed by NMS and box voting. Cascade R-CNN \cite{cai2018cascade} proposes a multi-stage object detection architecture which trains a sequence of R-CNN with increasing IoU thresholds. Thus the trained sequential R-CNN is sequentially more powerful for accurate localization during inference. RefineDet \cite{zhang2018single_RefineDet} improves the localization accuracy of the single-stage detector by using two-step bounding box regression. The anchor refinement module(ARM) firstly refines the human-designed anchors to improve the localization accuracy of human-designed anchors, then the object detection module(ODM) uses these more accurate anchors for the second step bounding box regression to improve the localization accuracy of the final detections. Libra R-CNN \cite{pang2019libraRCNN} designs balanced L1 loss to promote the regression gradients from inliers(accurate samples) during training. Thus, the trained regression branch is more powerful for accurate localization. Similarly, the IoU-balanced localization loss \cite{2019IoU-balancedlosses} reweights the localization loss for each positive example based on their regressed IoU. This reweighting procedure can down-weight the gradients from outliers and up-weight the gradients from inliers, thus improving the localization accuracy of models. Differently, the IoU-aware single-stage object detector improves the localization accuracy by predicting the localization accuracy for each detection and suppressing the detections of low localization accuracy based on the computed detection confidence during NMS and AP computation.

\textbf{Anchor-free single-stage object detectors}. To overcome the drawbacks of anchor-based detector, anchor-free single-stage object detectors have become more and more popular. Densebox\cite{huang2015densebox} proposes a single FCN that directly predicts bounding boxes and object confidences at every pixel of the feature map without using predifined anchors. The predicted box is represented by 4 distances between the current pixel and the four bounds of the predicted box. Unitbox\cite{yu2016unitbox} claims that the 4-D distance vector representing the predicted box should be optimized jointly as a whole, thus IoU loss is proposed to repalce the L2 loss for optimizing the predicted box. FCOS \cite{2019FCOS} solves object detection in a per-pixel prediction fashion based on a FCN. It consists of three prediction heads: classification head used for classification, regression head used for localization, centerness head used for predicting the centerness of each detected box. During inference, the predicted centerness of each detected box is multiplied by the corresponding classification score as the final score, which is used in the subsequent NMS and AP computation to suppress the poorly localized detections. PolarMask \cite{2019PolarMask} modifies FCOS to realize the instance segmentation. Similarly, centerness head is also used to suppress the segmentations of low localization accuracy and improve the localization accuracy of the model. The IoU-aware single-stage object detector designs an IoU prediction head parallel with the regression head to predict the IoU of each detection and the predicted IoU can be used to suppress the poorly localized detections. Differently, the IoU-aware single-stage object detector is a anchor-based detector and the IoU of each detected box is predicted.

\section{Method}
\label{S:3}
In this section, we introduce the model architecture of the IoU-aware single-stage object detector and different designing choices in details.

\subsection{IoU-aware single-stage object detector}
\label{subS:3.1}

\begin{figure}[h]
\centering
\includegraphics[height=0.34\linewidth]{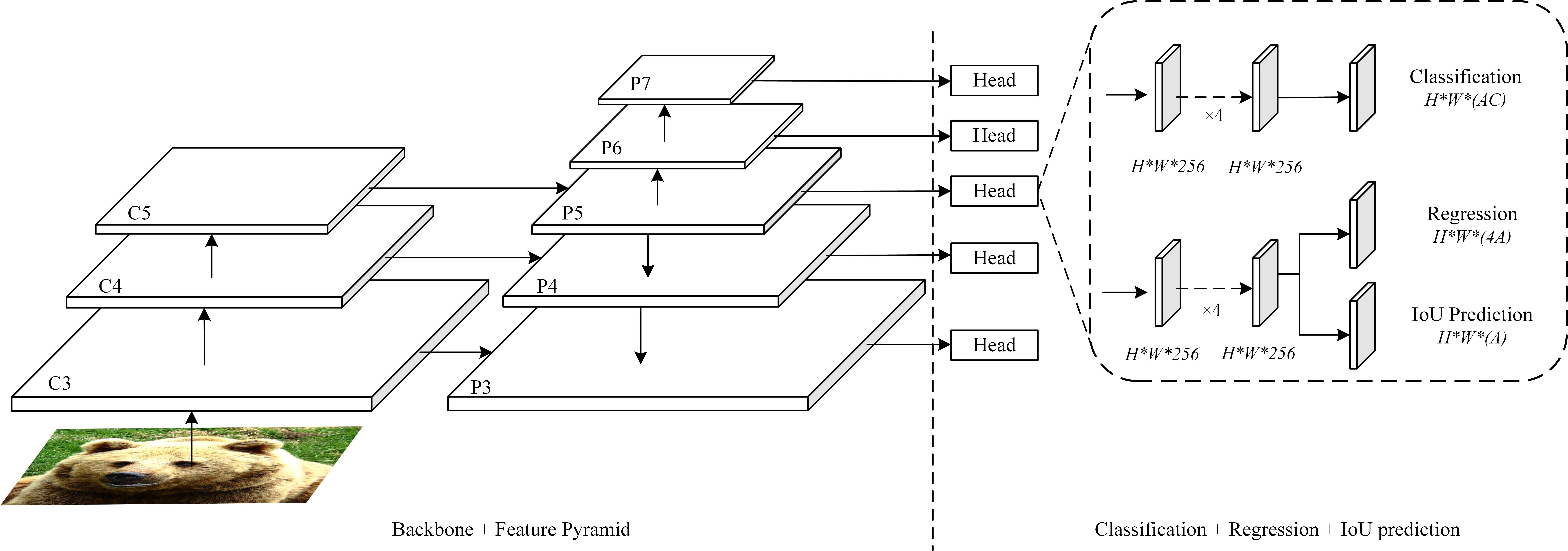}
\caption{The model architecture of IoU-aware single-stage object detector. The same backbone and feature pyramid network(FPN) are adopted as RetinaNet. An IoU prediction head is designed parallel with the regression head at the last layer of regression branch to predict the IoU for each detected box. The classification head, regression head and IoU prediction head all consist of only a single 3*3 convolution layer.}
\label{fig:architecture}
\end{figure}

The IoU-aware single-stage object detector is mostly modified from RetinaNet \cite{lin2017focal} with the same backbone and feature pyramid network(FPN) as Fig.\ref{fig:architecture} shows. Different from the RetinaNet, an IoU prediction head parallel with the regression head is designed in the last layer of regression branch to predict the IoU for each detected box while the classification branch is kept the same. To keep the model's efficiency, the IoU prediction head consists of only a single 3*3 convolution layer, followed by a sigmoid activation layer to ensure that the predicted IoU is in the range of [0, 1]. There are many other choices about the design of the IoU prediction head, such as designing an independent IoU prediction branch being the same as the classification branch and regression branch, but this kind of design will severely hurt the model's efficiency. Our design brings negligible computation burden to the whole model and can still substantially improve the model's performance.

\subsection{Training}
\label{subS:3.2}
\begin{figure}[h]
\centering
\includegraphics[height=0.34\linewidth]{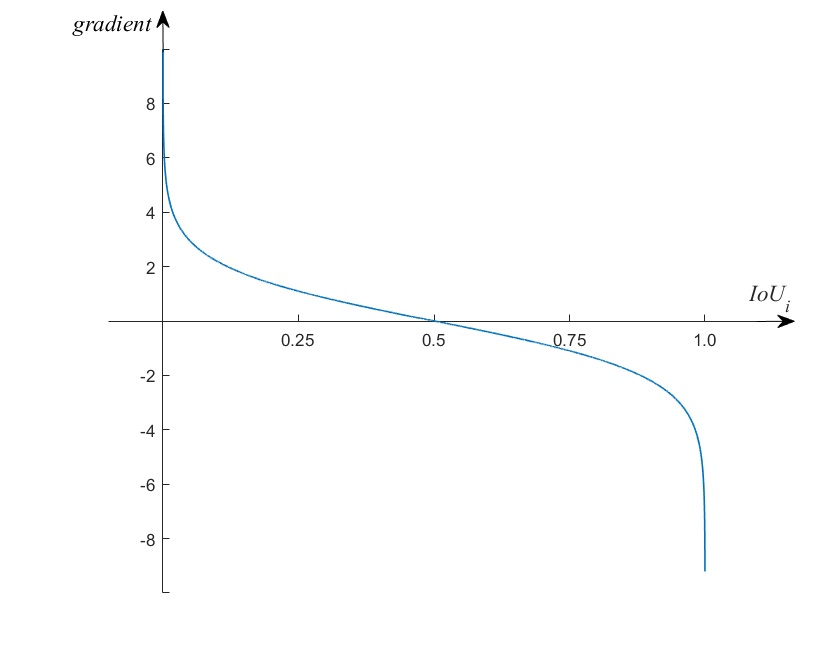}
\caption{The gradient of IoU prediction loss($L_{IoU}$) with respect to the target IoU($\hat{IoU}_i$).}
\label{fig:BCE_gradient}
\end{figure}

As the same as RetinaNet, the focal loss is adopted for the classification loss and the smooth L1 loss is adopted for the regression loss as Equ.\ref{eq1},\ref{eq2} show. The binary cross-entropy loss(BCE) is adopted for the IoU prediction loss and only the losses for the positive examples are computed as shown in th Equ.\ref{eq3}. $IoU_i$ represents the predicted IoU for each detected box and $\hat{IoU}_i$ is the target IoU computed between the regressed positive example $b_i$ and the corresponding ground truth box $\hat{b}_i$ as shown in Equ.\ref{eq3-1}. During training, whether to compute the gradient of $L_{IoU}$ with respect to $\hat{IoU}_i$ makes difference to the model's performance. This is caused by that the gradient from IoU prediction head can be back-propagated to the regression head if the gradient of $L_{IoU}$ with respect to $\hat{IoU}_i$ is computed during training. The gradient is computed as shown in Equ.\ref{eq3-2} and plot in Fig.\ref{fig:BCE_gradient}. Two observations can be obtained. Firstly, because the predicted IoU for most of the positive examples is not smaller than 0.5, the gradient is mostly non-positive and will guide the regression head to predict box $b_i$ that increases the target IoU($\hat{IoU}_i$) which is computed between the predicted box $b_i$ and the corresponding ground truth box $\hat{b}_i$. Secondly, as the predicted IoU(${IoU}_i$) increases, the magnitude of gradient that increases the target IoU($\hat{IoU}_i$) increases. This reduces the gap between the predicted IoU(${IoU}_i$) and the target IoU($\hat{IoU}_i$) and makes the predicted IoU more correlated with the target IoU. These two effects make our method more powerful for accurate localization as demonstrated in the following experiments. Other kinds of loss functions can also be considered, such as L2 loss and L1 loss. These different loss functions are compared in the following experiments. During training, the IoU prediction head is trained jointly with the classification head and regression head.

\begin{equation}
\label{eq1}
{{L}_{cls}}=\frac{1}{N_{Pos}}(\sum\limits_{i\in Pos}^{N}{\operatorname{FL}({{p}_{i}},{{{\hat{p}}}_{i}})+\sum\limits_{i\in Neg}^{M}{\operatorname{FL}({{p}_{i}},{{{\hat{p}}}_{i}})}})
\end{equation}

\begin{equation}
\label{eq2}
    {{L}_{loc}}=\frac{1}{N_{Pos}}\sum\limits_{i\in Pos}^{N}{\sum\limits_{m\in cx,cy,w,h}{{\text{smoot}{{\text{h}}_{\text{L1}}}(l_{i}^{m}-\hat{g}_{i}^{m})}}}
\end{equation}

\begin{equation}
\label{eq3}
{{L}_{IoU}}=\frac{1}{N_{Pos}}\sum\limits_{i\in Pos}^{N}{\operatorname{BCE}({{IoU}_{i}},{{{\hat{IoU}}}_{i}})}
\end{equation}

\begin{equation}
\label{eq3-1}
\hat{IoU_i}=\textup{overlap}(b_i,\hat{b_i})
\end{equation}

\begin{equation}
\label{eq3-2}
\frac{\partial \textup{BCE}(IoU_i,\hat{IoU_i})}{\partial \hat{IoU_i}}=\log\frac{1-IoU_i}{IoU_i}
\end{equation}

\begin{equation}
\label{eq4}
{{L}_{total}}={{L}_{cls}} + {{L}_{loc}} + {{L}_{IoU}}
\end{equation}

\subsection{Inference}
\label{subS:3.3}
During inference, the classification score $p_i$ is multiplied by the predicted IoU $IoU_i$ for each detected box to calculate the final detection confidence $S_{det}$ as Equ.\ref{eq5} shows. The parameter $\alpha$ in the range of [0, 1] is designed to control the contribution of the classification score and predicted IoU to the final detection confidence. This detection confidence can simultaneously be aware of the classification score and localization accuracy and thus is more correlated with the localization accuracy than the classification score only. And it is used to rank all the detections in the subsequent NMS and AP computation. The rankings of poorly localized detections with high classification score decrease while the rankings of well localized detections with low classification score increase, thus improving the localization accuracy of the models.

\begin{equation}
\label{eq5}
S_{det} = p_{i}^{\alpha}IoU_{i}^{(1-\alpha)}
\end{equation}

\section{Experiments}
\label{S:4}
\subsection{Experimental Settings}

\textbf{Dataset and Evaluation Metrics}. Most of the experiments are evaluated on the challenging MS COCO \cite{lin2014microCOCO} dataset. It consists of 118k images for training (\textit{train-2017}), 5k images for validation (\textit{val-2017}) and 20k images with no disclosed labels for test (\textit{test-dev}). There exist totally over 500k annotated object instances from 80 categories in the dataset. To demonstrate the generalization ability of our method, we also conduct experiments on the PASCAL VOC \cite{everingham2010pascalVOC} dataset in the ablation studies. VOC2007 consists of 5011 images for training (\textit{VOC2007 trainval}) and 4952 images for test (\textit{VOC2007 test}). And VOC2012 consists of 17125 images for training (\textit{VOC2012 trainval}) and 5138 images for test (\textit{VOC2012 test}). For all the experiments, the standard COCO-style Average Precision (AP) metrics are adopted which consist of  AP (averaged AP at IoUs from 0.5 to 0.95 with an interval of 0.05), $\text{A}{{\text{P}}_{50}}$ (AP at IoU threshold 0.5), $\text{A}{{\text{P}}_{75}}$ (AP at IoU threshold 0.75), $\text{A}{{\text{P}}_{S}}$ (AP for objects of small scales), $\text{A}{{\text{P}}_{M}}$ (AP for objects of medium scales) and $\text{A}{{\text{P}}_{L}}$ (AP for objects of large scales).

\textbf{Implementation Details}. All the object detection models are implemented based on PyTorch \cite{paszke2019pytorch} and MMDetection \cite{chen2019mmdetection}. As only 2 GPUs are available, linear scaling rule \cite{goyal2017accurate} is adopted to adjust the learning rate during training. For the main results, all the models are evaluated on COCO \textit{test-dev}. The converged models provided by MMDetection are evaluated as the baselines. With the default setting in the MMDetection, the IoU-aware single-stage object detectors are all trained for total 12 epochs with the image scale of [800, 1333]. Some papers report the main results obtained by training the models with total 1.5 longer time and scale jitter. These tricks are not adopted in our experiments. In the ablation studies, the IoU-aware single-stage object detector with ResNet50 as backbone is trained on COCO \textit{train-2017} and evaluated on COCO \textit{val-2017} using the image scale of [600, 1000]. For the experiments on PASCAL VOC, the models with different backbones are trained on the \textit{VOC2007 trainval} and \textit{VOC2012 trainval} and evaluated on \textit{VOC2007 test} with the image scale of [600, 1000]. If not specified, all the other settings are kept the same as the default settings in the MMDdetection.

\begin{table}[h]
\centering
\resizebox{1\linewidth}{!}{

\begin{tabular}{l l l l l l l l l}
\hline
{Model} & {Backbone} & {Schedule} & {$\text{AP}$} & {$\text{AP}_{50}$} & {$\text{AP}_{75}$} & {$\text{AP}_S$} & {$\text{AP}_M$} & {$\text{AP}_L$}\\
\hline
YOLOv2 \cite{redmon2017yolo9000} & DarkNet-19 & - & 21.6 & 44.0 & 19.2 & 5.0  & 22.4 & 35.5 \\
YOLOv3 \cite{redmon2018yolov3}   & DarkNet-53 & - & 33.0 & 57.9 & 34.4 & 18.3 & 35.4 & 41.9 \\
SSD300 \cite{liu2016ssd}         & VGG16      & - & 23.2 & 41.2 & 23.4 & 5.3  & 23.2 & 39.6 \\
SSD512 \cite{liu2016ssd}         & VGG16      & - & 26.8 & 46.5 & 27.8 & 9.0  & 28.9 & 41.9 \\
Faster R-CNN \cite{ren2015faster}& ResNet-101-FPN & - & 36.2 & 59.1 & 39.0 & 18.2 & 39.0 & 48.2 \\
Deformable R-FCN \cite{dai2017deformableConv} & Inception-ResNet-v2 & - & 37.5 & 58.0 & 40.8 & 19.4 & 40.1 & 52.5 \\
Mask R-CNN \cite{he2017mask} & ResNet-101-FPN & - & 38.2 & 60.3 & 41.7 & 20.1 & 41.1 & 50.2 \\
\hline
Faster R-CNN* & ResNet-50-FPN & 1x & 36.2 & 58.5 & 38.9 & 21.0 & 38.9 & 45.3 \\
Faster R-CNN* & ResNet-101-FPN & 1x & 38.8 & 60.9 & 42.1 & 22.6 & 42.4 & 48.5 \\
Faster R-CNN* & ResNeXt-32x8d-101-FPN & 1x & 40.3 & 62.7 & 44.0 & 24.4 & 43.7 & 49.8 \\
RetinaNet* & ResNet-50-FPN & 1x & 35.9 & 55.8 & 38.4 & 19.9 & 38.8 & 45.0 \\
RetinaNet* & ResNet-101-FPN & 1x & 38.1 & 58.5 & 40.8 & 21.2 & 41.5 & 48.2 \\
RetinaNet* & ResNeXt-32x8d-101-FPN & 1x & 39.4 & 60.2 & 42.3 & 22.5 & 42.8 & 49.8 \\
\hline
IoU-aware RetinaNet & ResNet-50-FPN & 1x & 36.9 & 56.1 & 40.1 & 20.9 & 40.0 & 46.0 \\
IoU-aware RetinaNet & ResNet-101-FPN & 1x & 39.2 & 58.2 & 42.9 & 22.1 & 42.7 & 50.0 \\
IoU-aware RetinaNet & ResNeXt-32x8d-101-FPN & 1x & 40.6 & 60.1 & 44.2 & 23.4 & 43.9 & 51.8\\
IoU-aware RetinaNet$\dagger$ & ResNet-50-FPN & 1x & 37.8 & 55.1 & 40.9 & 21.1 & 41.2 & 47.3 \\
IoU-aware RetinaNet$\dagger$ & ResNet-101-FPN & 1x & 39.7 & 56.2 & 43.3 & 21.9 & 43.4 & 51.6 \\
IoU-aware RetinaNet$\dagger$ & ResNeXt-32x8d-101-FPN & 1x & 41.1 & 58.7 & 44.5 & 23.3 & 44.8 & 52.7\\

\hline
\end{tabular}
}
\caption{ Comparison with the state-of-the-art methods on COCO \textit{test-dev}. The symbol "*" means the reimplementation results in MMDetection \cite{chen2019mmdetection}. The symbol "$\dagger$" means the gradient of $L_{IoU}$ with respective to ${\hat{IoU}}_{i}$ is computed during training. The training schedule is the same as Detectron \cite{Detectron2018}. "1x" means the model is trained for total 12 epochs. Different from some research, the longer training schedule and scale jitters are not adopted in our experiments.}
\label{table:MainResults}
\end{table}

\subsection{Main Results}

In the main results as shown in Table~\ref{table:MainResults}, the performance of the IoU-aware single-stage object detectors with different backbones are compared with the state-of-the-art object detection models on the COCO \textit{test-dev}. For fair comparison, the trained models provided by MMDetectioin \cite{chen2019mmdetection} with different backbones are evaluated as the baselines. As Table \ref{table:MainResults} shows, the IoU-aware RetinaNets with different backbones can substantially improve AP by $1.7\%\sim1.9\%$ compared with the baselines. In addition, the performance for AP75 is largely improved by $2.2\%\sim2.5\%$ while the performance for AP50 is decreased by $0.7\%\sim2.3\%$, which demonstrates the effectiveness of the IoU-aware RetinaNet on improving the models' localization accuracy. In addition, the performance of IoU-aware RetinaNets has surpassed that of the two-stage detector Faster R-CNN with the same backbone by $0.8\%\sim1.6\%$ AP and the improvement mostly comes from the high localization accuracy of the IoU-aware RetinaNets.

\subsection{Ablation Studies}

\begin{table}[h]
\centering
%\resizebox{1\linewidth}{!}{
\begin{tabular}{l l l l l l l l}
\hline
{IoU prediction loss} & {$\text{AP}$} & {$\text{AP}_{50}$} & {$\text{AP}_{75}$} & {$\text{AP}_S$} & {$\text{AP}_M$} & {$\text{AP}_L$}\\
\hline
baseline   &34.4 &54.1 &36.6 &17.3 &38.5 &47.1 \\
L2 loss    &35.1 &53.8 &37.9 &18.6 &39.2 &47.6 \\
BCE        &35.4 &54.1 &38.2 &18.9 &39.3 &48.3 \\
\hline
\end{tabular}
%}
\caption{ The effectiveness of training IoU-aware RetinaNet-ResNet50 with different IoU prediction losses on COCO \textit{val-2017}.}
\label{table:IoUPredictionLoss}
\end{table}

\textbf{IoU Prediction Loss}. Different IoU prediction losses are used to train the IoU-aware RetinaNet. To investigate the effect of IoU prediction loss only, the detection confidence is computed by multiplying the classification score and predicted IoU directly without using the parameter $\alpha$. As shown in Table \ref{table:IoUPredictionLoss}, training the model with binary cross-entropy loss can produce better performance than training the model with L2 loss. Thus binary cross-entropy loss is adopted in all the subsequent experiments.

\begin{table}[h]
\centering
\resizebox{1\linewidth}{!}{
\begin{tabular}{l l l l l l l l l l l l}
\hline
{$\alpha$} & {$\text{AP}$} & {$\text{AP}_{50}$} & {$\text{AP}_{75}$} & {$\text{AP}_S$} & {$\text{AP}_M$} & {$\text{AP}_L$}& {$\text{AP}_{60}$}& {$\text{AP}_{70}$}& {$\text{AP}_{80}$} & {$\text{AP}_{90}$}\\
\hline
baseline   &34.4 &54.1 &36.6 &17.3 &38.5 &47.1 &49.6 &41.8 &29.9 &10.9\\
none       &35.4 &54.1 &38.2 &18.9 &39.3 &48.3 &50.0 &43.3 &31.9 &12.2\\
1.0        &34.5 &54.1 &36.7 &17.4 &38.4 &46.8 &49.5 &42.0 &30.1 &11.0\\
0.9        &34.8 &54.3 &37.1 &17.8 &38.6 &47.0 &49.7 &42.4 &30.4 &11.2\\
0.8        &35.0 &54.4 &37.4 &18.3 &38.8 &47.6 &49.9 &42.6 &30.7 &11.4\\
0.7        &35.2 &54.4 &37.7 &18.5 &39.0 &47.9 &50.1 &43.0 &31.2 &11.7\\
0.6        &35.4 &54.3 &38.0 &18.8 &39.2 &48.2 &50.2 &43.2 &31.5 &12.0\\
0.5        &35.5 &54.1 &38.2 &18.9 &39.4 &48.3 &50.0 &43.3 &31.9 &12.2\\
0.4        &35.5 &53.7 &38.4 &18.9 &39.5 &48.4 &49.8 &43.3 &32.5 &12.5\\
0.3        &35.4 &53.0 &38.5 &18.8 &39.5 &48.4 &49.3 &43.2 &32.8 &12.8\\
\hline
\end{tabular}
}
\caption{ The effectiveness of not computing the gradient of $L_{IoU}$ with respective to ${\hat{IoU}}_{i}$ during training on COCO \textit{val-2017}.}
\label{table:NotComputingGradientOftruthIoU}
\end{table}

\textbf{Detection Confidence Computation}. During inference, the detection confidence is computed according to Equ.~\ref{eq5} and the parameter $\alpha$ is used to control the contribution of the  classification score and predicted IoU to the final detection confidence. In addition, the detection confidence can also be computed by multiplying the classification score and predicted IoU directly without using the parameter $\alpha$. There are several observations from the experimental results as Table \ref{table:NotComputingGradientOftruthIoU} shows. Firstly,  multiplying the classification score and predicted IoU with the parameter $\alpha$ equaling to 0.5 can improve AP by $1.1\%$, which is slightly better than computing the detection confidence without using $\alpha$. Thus computing the detection confidence using the parameter $\alpha$ as Equ. \ref{eq5} shows is used in our paper. Secondly, when $\alpha$ equals to 1.0, only the classification score is used as the detection confidence and the AP is improved by $0.1\%$. This demonstrates that multi-task training with IoU prediction loss is beneficial to the model's performance. Thirdly, when $\alpha$ equals to 0.4 or 0.5, the best performance of AP $35.5\%$ is obtained, which is $1.1\%$ better than the baseline. The AP50 marginally decreases by $0\%\sim0.4\%$ while the $\text{AP70}\sim\text{AP90}$ increase by $1.3\%\sim2.6\%$, demonstrating the effectiveness of our method on improving the model's localization accuracy. Finally, with the decrease of the parameter $\alpha$ value, the contribution of the predicted IoU to the detection confidence is improved and the performance at higher IoU threshold is improved, which demonstrates that the predicted IoU is more correlated with the localization accuracy and can bias the model to the detections with high localization accuracy. Thus lower $\alpha$ can be selected for the application where model's localization accuracy is more important.

\begin{table}[h]
\centering
\resizebox{1\linewidth}{!}{
\begin{tabular}{l l l l l l l l l l l l}
\hline
{$\alpha$} & {$\text{AP}$} & {$\text{AP}_{50}$} & {$\text{AP}_{75}$} & {$\text{AP}_S$} & {$\text{AP}_M$} & {$\text{AP}_L$}& {$\text{AP}_{60}$}& {$\text{AP}_{70}$}& {$\text{AP}_{80}$} & {$\text{AP}_{90}$}\\
\hline
baseline   &34.4 &54.1 &36.6 &17.3 &38.5 &47.1 &49.6 &41.8 &29.9 &10.9\\
none       &36.0 &52.7 &39.0 &18.4 &40.1 &50.0 &49.0 &43.0 &33.3 &15.9\\
1.0        &34.9 &52.2 &37.4 &16.7 &39.4 &48.8 &48.2 &41.7 &31.7 &14.6\\
0.9        &35.4 &52.9 &37.9 &17.5 &39.6 &49.1 &48.8 &42.3 &32.1 &14.8\\
0.8	       &35.7 &53.1 &38.3 &18.0 &39.8 &49.3 &49.1 &42.6 &32.4 &15.1\\
0.7	       &35.8 &53.1 &38.5 &18.1 &39.9 &49.5 &49.1 &42.8 &32.7 &15.3\\
0.6	       &35.9 &53.0 &38.7 &18.3 &40.0 &49.8 &49.1 &42.9 &33.0 &15.6\\
0.5	       &36.0 &52.8 &39.0 &18.4 &40.2 &50.0 &49.0 &43.0 &33.3 &15.9\\
0.4	       &36.1 &52.4 &39.2 &18.4 &40.2 &50.2 &48.8 &43.0 &33.6 &16.3\\
0.3	       &35.9 &51.6 &39.2 &18.2 &40.1 &50.2 &48.2 &43.0 &33.9 &16.6\\
\hline
\end{tabular}
}
\caption{ The effectiveness of computing the gradient of $L_{IoU}$ with respective to ${\hat{IoU}}_{i}$ during training on COCO \textit{val-2017}.}
\label{table:gradient of IoU truth}
\end{table}

\textbf{The Effectiveness of Computing the Gradient of $L_{IoU}$ with Respective to ${{{\hat{IoU}}}_{i}}$ During Training.} All the experimental results above are obtained by training the models without computing the gradient of $L_{IoU}$ with respective to ${\hat{IoU}}_{i}$ during training. Because the ground truth IoU (${\hat{IoU}}_{i}$) is computed between the predicted box and the corresponding ground truth box for each positive example, the gradient from the IoU prediction head will be back-propagated to the regression head if the gradient of $L_{IoU}$ with respective to ${\hat{IoU}}_{i}$ is computed during training. Several observations can be drawn from the experimental results shown in Table~\ref{table:gradient of IoU truth}. Firstly, computing the gradient of $L_{IoU}$ with respective to ${\hat{IoU}}_{i}$ during training can improve AP by $0.6\%$ compared with not computing the gradients. Thus it is selected in our paper. Secondly, when the parameter $\alpha$ equals to 1.0, the AP is improved by $0.5\%$ compared with the baseline. In addition, \text{AP80} and \text{AP90} increase by $1.8\%\sim3.7\%$ while \text{AP50} and \text{AP60} decrease by $1.4\%\sim1.9\%$ which means that the gradients from the IoU prediction head make the regression head more powerful for accurate localization. Finally, when $\alpha$ equals to 0.4, the IoU-aware RetinaNet can substantially improve AP, AP80, AP90 by $1.7\%$, $3.7\%$, $5.4\%$ respectively compared with the baseline, demonstrating the powerful capability of our method for accurate localization.

\begin{table}[h]
\centering
\resizebox{1\linewidth}{!}{
\begin{tabular}{l l l l l l l l}
\hline
{Model} & {Backbone} & {$\text{AP}$} & {$\text{AP}_{50}$} & {$\text{AP}_{60}$} & {$\text{AP}_{70}$} & {$\text{AP}_{80}$} & {$\text{AP}_{90}$}\\
\hline
RetinaNet & ResNet-50-FPN         & 51.4 & 78.8 & 74.3 & 63.6 & 44.9 & 15.2 \\
RetinaNet & ResNet-101-FPN        & 55.1 & 81.1 & 77.2 & 67.5 & 50.4 & 20.1 \\
RetinaNet & ResNeXt-32x8d-101-FPN & 56.1 & 81.9 & 78.1 & 68.1 & 52.0 & 21.4 \\
\hline
IoU-aware RetinaNet & ResNet-50-FPN         & 53.6 & 79.0 & 75.1 & 66.1 & 48.7 & 19.4 \\
IoU-aware RetinaNet & ResNet-101-FPN        & 56.2 & 80.5 & 76.8 & 68.5 & 52.4 & 22.8 \\
IoU-aware RetinaNet & ResNeXt-32x8d-101-FPN & 57.6 & 81.6 & 78.0 & 69.5 & 54.7 & 24.7 \\
IoU-aware RetinaNet$\dagger$ & ResNet-50-FPN         & 55.8 & 79.5 & 75.4 & 67.0 & 51.6 & 25.4 \\
IoU-aware RetinaNet$\dagger$ & ResNet-101-FPN        & 58.0 & 80.1 & 76.9 & 68.8 & 55.0 & 29.3 \\
IoU-aware RetinaNet$\dagger$& ResNeXt-32x8d-101-FPN & 59.7 & 81.8 & 78.6 & 70.6 & 56.7 & 31.4 \\
\hline

\end{tabular}
}
\caption{Experimental results on PASCAL VOC. All the models are trained on \textit{VOC2007 trainval} and \textit{VOC2012 trainval} and evaluated on \textit{VOC2007 test} with the image scale of [600, 1000]. All the other settings are adopted as the same as the default settings provided in the MMDetection. The symbol "$\dagger $" means the gradient of $L_{IoU}$ with respective to ${\hat{IoU}}_{i}$ is computed during training.}
\label{table:VOC}
\end{table}

\textbf{Ablation Studies on PASCAL VOC}. As shown in Table \ref{table:VOC}, when the gradient of $L_{IoU}$ with respective to ${\hat{IoU}}_{i}$ is not computed during training, IoU-aware RetinaNets with different backbones can improve AP by $1.1\%\sim2.2\%$ compared with the baselines while the improvement for AP at higher IoU threshold(0.8,0.9) is $2.0\%\sim4.2\%$ , which demonstrates that our method can substantially improve the model's localization accuracy. When the gradient of $L_{IoU}$ with respective to ${\hat{IoU}}_{i}$ is computed during training, the AP is improved by $2.9\%\sim4.4\%$ while the AP at higher IoU threshold(0.8, 0.9) is improved by $4.6\%\sim10.2\%$. This demonstrates that computing the the gradient of $L_{IoU}$ with respective to ${\hat{IoU}}_{i}$ makes the IoU-aware RetinaNet more powerful especially for the accurate localization. The conclusions from the experimental results of PASCAL VOC dataset are consistent with those from the experimental results of COCO dataset, which demonstrates our method has generalization ability to other datasets and can be applied to different application scenes.

\begin{table}[h]
\centering
\resizebox{1\linewidth}{!}{
\begin{tabular}{l l l l l l l l l l}
\hline
{Backbone} & {$IoU_{pred}$} & {$IoU_{truth}$}& {$\alpha$} & {$\text{AP}$} & {$\text{AP}_{50}$} & {$\text{AP}_{75}$} & {$\text{AP}_S$} & {$\text{AP}_M$} & {$\text{AP}_L$}\\
\hline
ResNet-50-FPN         &        &        &     &35.6 &55.5 &38.3 &20.0 &39.6 &46.8 \\
ResNet-101-FPN        &        &        &     &37.7 &57.5 &40.4 &21.1 &42.2 &49.5 \\
ResNeXt-32x8d-101-FPN &        &        &     &39.0 &59.4 &41.7 &22.6 &43.4 &50.9 \\
\hline
ResNet-50-FPN         &$\surd$ &        &0.5  &37.3 &54.4 &40.2 &20.4 &41.2 &48.7 \\
ResNet-101-FPN        &$\surd$ &        &0.4  &39.4 &56.2 &42.9 &21.6 &44.0 &52.9 \\
ResNeXt-32x8d-101-FPN &$\surd$ &        &0.4  &40.9 &58.1 &44.3 &22.4 &45.7 &54.5 \\
\hline
ResNet-50-FPN         &        &$\surd$ &0.2  &50.1 &61.1 &57.7 &36.1 &56.9 &61.0 \\
ResNet-101-FPN        &        &$\surd$ &0.2  &52.1 &63.2 &59.7 &36.7 &59.1 &65.1 \\
ResNeXt-32x8d-101-FPN &        &$\surd$ &0.2  &53.2 &64.7 &60.8 &37.2 &60.3 &65.7 \\
\hline
\end{tabular}
}
\caption{ The performance gap of computing the detection confidence using the predicted IoU and ground truth IoU respectively on COCO \textit{val-2017} with image scale of [800, 1333]. The detection confidence is computed based on Equ.\ref{eq5} and the parameter $\alpha$ is adjusted to be optimal for computing the detection confidence using the predicted IoU and ground truth IoU respectively.}
\label{table:PerformanceUpperBound}
\end{table}

\subsection{Discussions}
\textbf{The Upper Bound of IoU-aware RetinaNet}. To evaluate the upper bound of IoU-aware RetinaNet, we replace the predicted IoU with the ground truth IoU for each detection to compute the detection confidence during inference. We define the ground truth IoU for each detection as the IoU between the detection and its' nearest ground truth box without considering categories. Specifically, the IoUs between each detection and all the ground truth boxes of all categories are computed in each image and then the maximal IoU is selected as the ground truth IoU of each detection, denoted as $IoU_{truth}$. As shown in Table~\ref{table:PerformanceUpperBound}, compared with RetinaNets, the IoU-aware RetinaNets with different backbones can improve AP by $1.7\%\sim1.9\%$ when using the predicted IoU but can improve AP by $14.2\%\sim14.5\%$ when using the ground truth IoU. There are still $12.3\%\sim12.8\%$ for AP to be improved for IoU-aware RetinaNet. From the observation of the large gap of the performance between using the predicted IoU and using the ground truth IoU, two meaningful conclusions can be drawn. Firstly, a huge number of objects have been successfully detected by the regressed boxes but suppressed or discarded during inference because of the low detection confidence. Secondly, although the predicted IoU of IoU-aware RetinaNet can alleviate the problem of mismatch between the detection confidence and localization accuracy, the predicted IoU is far from accurate compared with the ground truth IoU. If the accuracy of the predicted IoU can be improved further, a large improvement for the model's performance can be obtained. 

\begin{figure}[h]
\centering
\subfloat[]{\label{Fig:ConfidenceVSIoU1}%%
\includegraphics[width=0.5\linewidth]{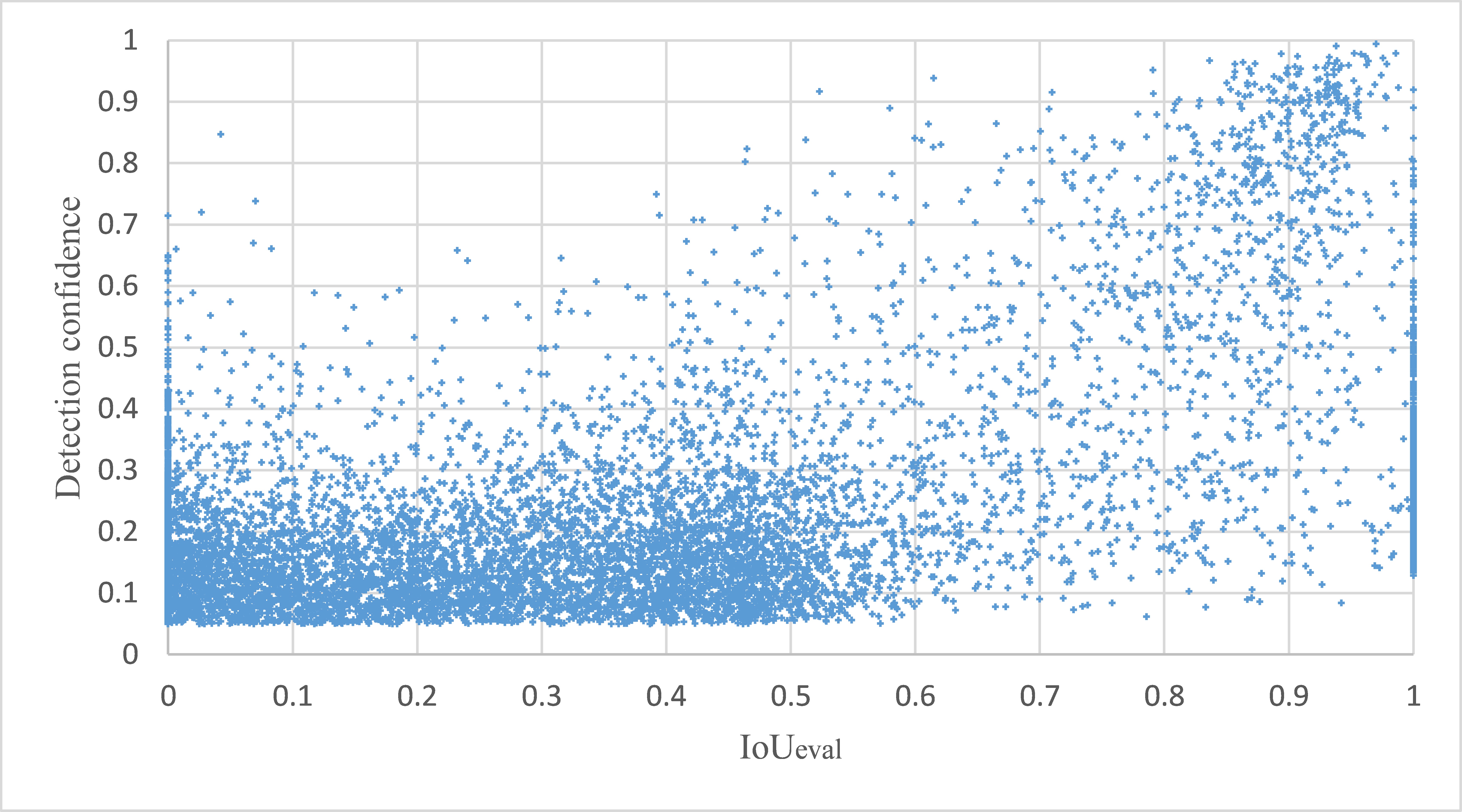}}
\subfloat[]{\label{Fig:ConfidenceVSIoU2}%%
\includegraphics[width=0.5\linewidth]{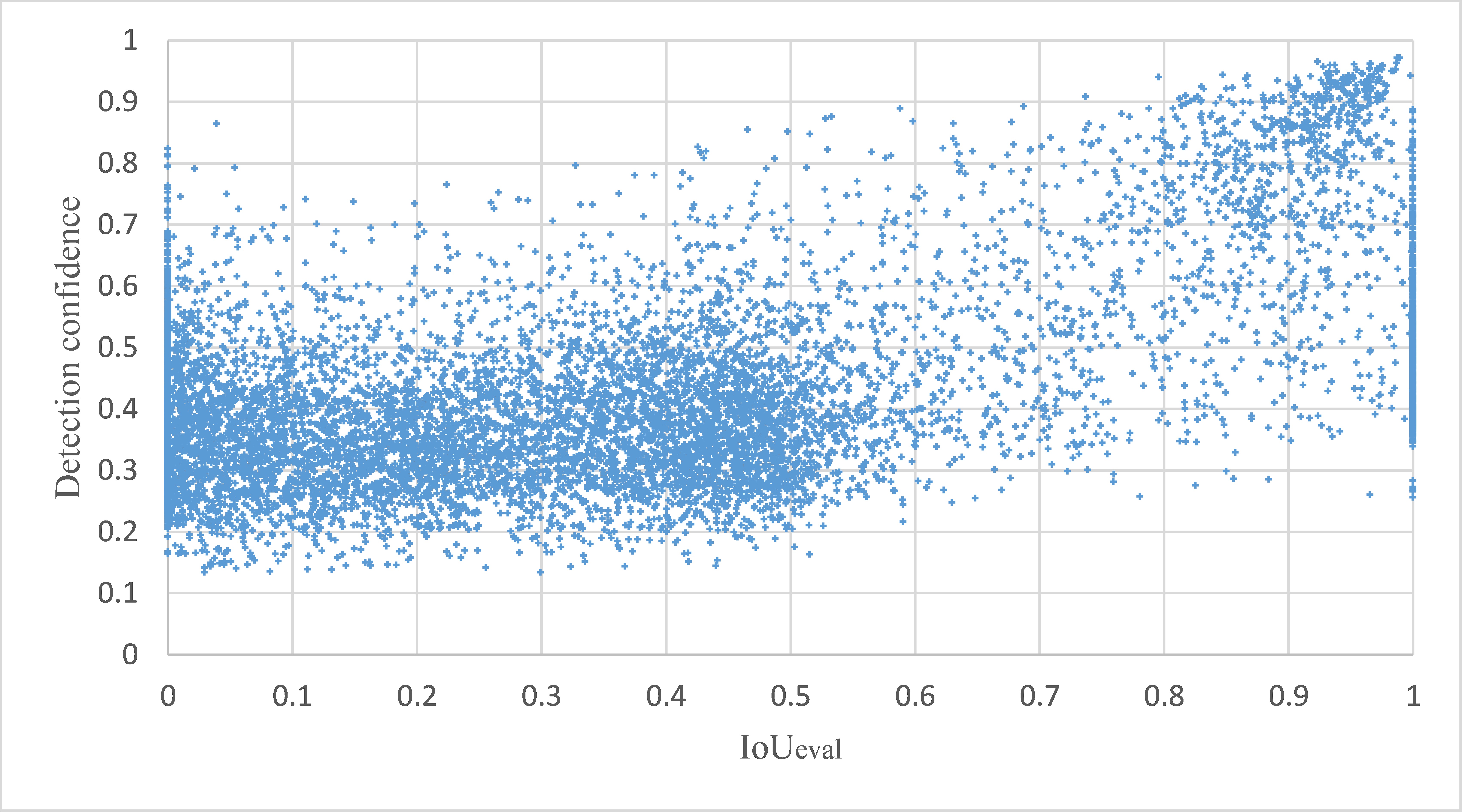}}

\subfloat[]{\label{Fig:ConfidenceVSIoU3}%%
\includegraphics[width=0.5\linewidth]{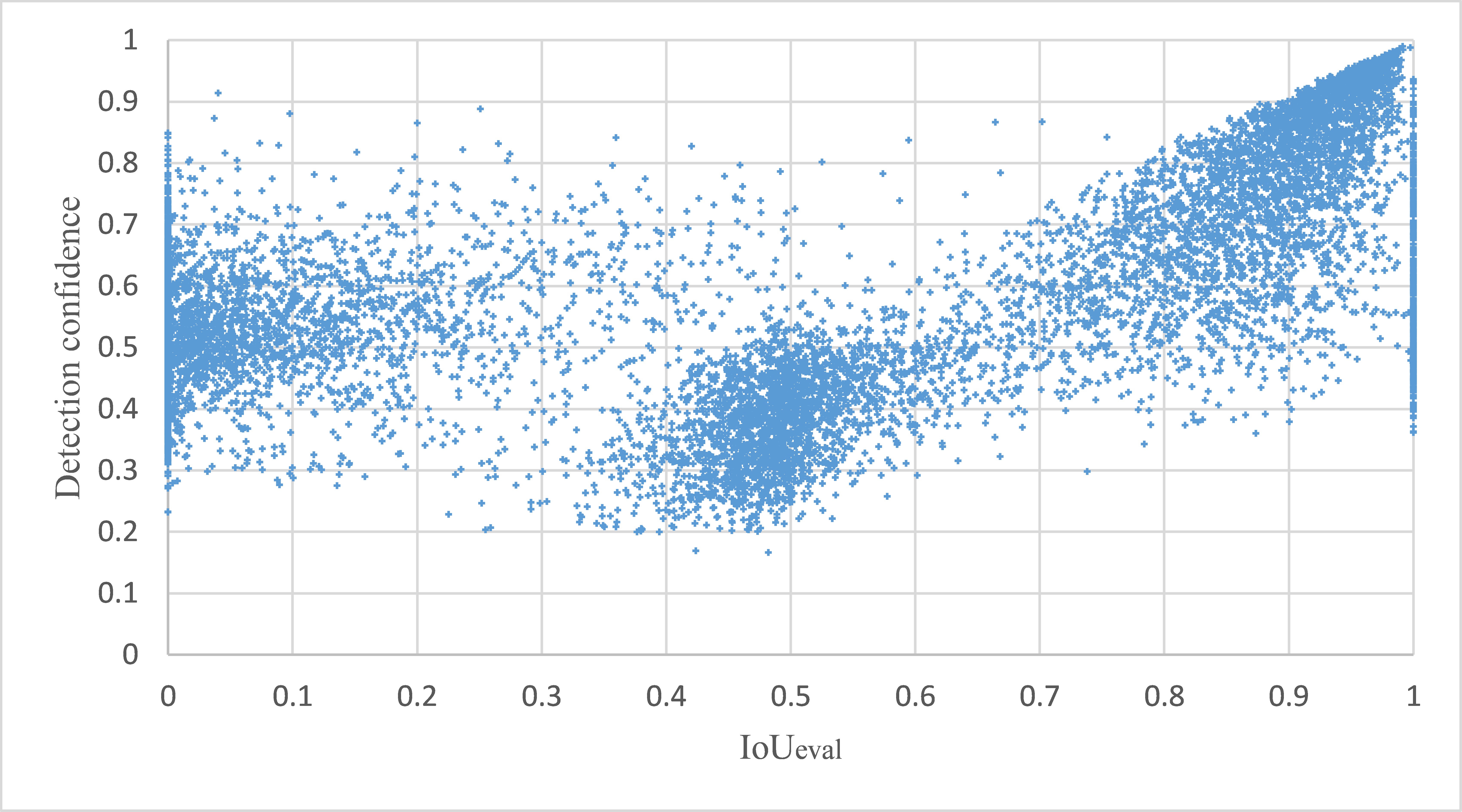}}
\caption{Detection confidence VS $IoU_{eval}$. 10K detections are sampled from (a) RetinaNet, (b) IoU-aware RetinaNet using the predicted IoU and (c) IoU-aware RetinaNet using the ground truth IoU respectively. For (a) RetinaNet, the detection confidence is the classification score. For (b) IoU-aware RetinaNet using the predicted IoU, the detection confidence is computed by multiplying the classification score and predicted IoU using Equ.\ref{eq5} with the optimal parameter $\alpha$. For (c) IoU-aware RetinaNet using the ground truth IoU, the detection confidence is computed by multiplying the classification score and ground truth IoU using Equ.\ref{eq5} with the optimal parameter $\alpha$.}
\label{fig:ConfidenceVSIoU}
\end{figure}

\textbf{Why Can IoU-aware RetinaNet Improve Model's Performance?} For clarity, we firstly define $IoU_{eval}$ used during evaluation which is different from $IoU_{truth}$ used during inference. During evaluating the detection results, the IoUs between each detection and all the ground truth boxes belonging to the same categories are computed and then the maximum IoU is selected to evaluate whether the detection is a truth positive example or a false positive example. We denoted this maximum IoU as $IoU_{eval}$. We select 10K detections from the baseline RetinaNet, IoU-aware RetinaNet using the predicted IoU and IoU-aware RetinaNet using the ground truth IoU respectively and plot the detections in the figures of detection confidence VS $IoU_{eval}$ respectively. As shown in Fig.\ref{Fig:ConfidenceVSIoU1}, there are a large number of detections from RetinaNet that have the high localization accuracy but low detection confidence. As Fig.\ref{Fig:ConfidenceVSIoU2} shows, IoU-aware RetinaNet using the predicted IoU can slightly increase the detection confidence of this kind of detections and the mismatch problem between the detection confidence and localization accuracy is alleviated slightly compared with RetinaNet. This is the reason why IoU-aware RetinaNet can improve model's performance especially the localization accuracy. When computing the detection confidence using the ground truth IoU during inference, the detection confidence becomes strongly correlated with the localization accuracy as Fig.\ref{Fig:ConfidenceVSIoU3} shows. Comparison between Fig.\ref{Fig:ConfidenceVSIoU2} and Fig.\ref{Fig:ConfidenceVSIoU3} also shows that there still exists a large gap between the predicted IoU and ground truth IoU which leaving a large improvement room for the model's performance and more research needs to be done to improve the accuracy of the predicted IoU.

\begin{figure}[h]
\centering
\subfloat[]{\label{Fig:IoUtruth_IoUeval1}%%
\includegraphics[width=0.5\linewidth]{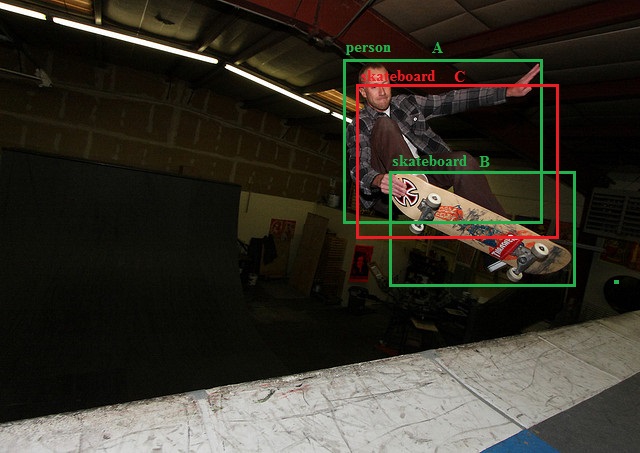}}
\subfloat[]{\label{Fig:IoUtruth_IoUeval2}%%
\includegraphics[width=0.5\linewidth]{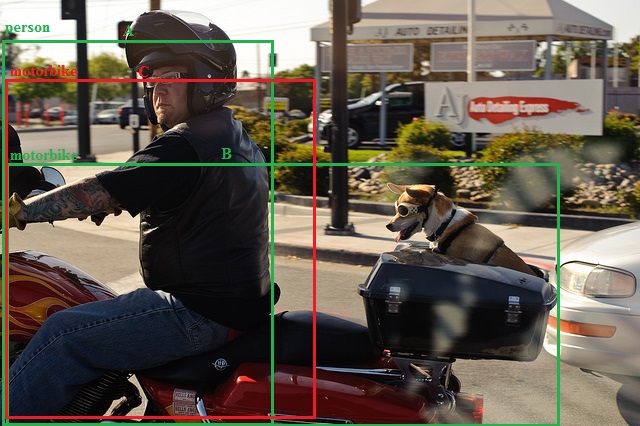}}
\caption{Illustration for the definition of $IoU_{truth}$ and $IoU_{eval}$. (a) $IoU_{truth}=IoU_{AC}=0.7$ and $IoU_{eval}=IoU_{BC}=0.3$. The green boxes A and B are the ground truth boxes and the red box C is the detected box. During inference, IoUs between the detected box and ground truth boxes are computed without considering categories and the green box A is the nearest ground box for the detected box C, thus $IoU_{AC}=0.7$ is defined as the ground truth IoU for the detected box C. During evaluating the detection results, IoUs between the detected box and the ground truth boxes belonging to the same category are computed and the green box B is the nearest ground truth box belonging to the same category for the detected box C. Thus $IoU_{BC}=0.3$ is defined as the $IoU_{eval}$ for the detected box C. (b) $IoU_{truth}=IoU_{AC}=0.8$ and $IoU_{eval}=IoU_{BC}=0.5$}
\label{fig:IoUtruth_IoUeval}
\end{figure}

\textbf{The Error of Classification.} Interestingly, the detections with $IoU_{eval}$ in the interval of [0, 0.3] have relatively high detection confidence as shown in Fig.\ref{Fig:ConfidenceVSIoU3}, meaning that the $IoU_{truth}$ of these detections computed during inference is large while their $IoU_{eval}$ computed during evaluation is small in the range of [0, 0.3]. This kind of detections can be considered as the misclassified detections. As shown in Fig.\ref{Fig:IoUtruth_IoUeval1}, the green boxes A and B are ground truth boxes while the red box C is the detected box. Because $IoU_{truth}=IoU_{AC}=0.7$ and $IoU_{eval}=IoU_{BC}=0.3$, the person can be considered having been detected by the red box if classification is not considered. However, the classification head makes the wrong prediction about the red detected box. The detections with $IoU_{eval}$ in the interval of [0, 0.3] as shown in Fig.\ref{Fig:ConfidenceVSIoU3} are all this kind of detections which are misclassified by the classification head. Feature alignment such as RoIConv \cite{chen2019FeaturAlign} may be helpful for solving this misclassification problem.

\section{Conclusions}
\label{S:5}

In this work, we demonstrate that the low correlation between the classification score and localization accuracy of the single-stage object detector can severely hurt the localization accuracy of models. Thus, IoU-aware single-stage object detector is designed by adding an IoU prediction head at the last layer of the regression branch to predict the IoU of each detected box. In this way, the model is aware of the localization accuracy of each detection. During inference, the detection confidence is computed by multiplying the classification score and predicted IoU and then used to rank all the detections in the subsequent NMS and AP computation. Extensive experiments on MS COCO dataset and PASCAL VOC dataset have shown that IoU-aware single-stage object detectors can substantially improve the model's performance, especially the localization accuracy. In addition, we demonstrate that there still exists a large gap between the predicted IoU and the ground truth IoU which substantially limits the performance of our method. Feature alignment and attention mechanism may be important for the accurate IoU prediction and this will be left for the future research.

\section{Acknowledgements}
\label{S:6}
This research did not receive any specific grant from funding agencies in the public, commercial, or not-for-profit sectors.



%% References



%% New version of the num-names style
\bibliographystyle{elsarticle-num-names}
\bibliography{sample.bib}






% \end{thebibliography}

\end{document}